\documentclass[sigconf]{acmart}
\AtBeginDocument{%
  }
\renewcommand\footnotetextcopyrightpermission[1]{}
\setcopyright{none}
\usepackage[ruled,vlined]{algorithm2e}  
\usepackage{enumitem}
\usepackage{booktabs}
\usepackage{graphicx}

\usepackage{multirow}
\usepackage{xcolor}
\usepackage{array}
\usepackage{fontawesome5}
\usepackage{hyperref}
\renewcommand{\arraystretch}{1.1}
\usepackage{xcolor}
\usepackage{hyperref}
\usepackage[most]{tcolorbox}

\newtcolorbox{promptbox}[1]{
  colback=gray!6,
  colframe=gray!55,
  title=#1,
  fonttitle=\bfseries,
  boxrule=0.6pt,
  arc=2pt,
  left=6pt,
  right=6pt,
  top=6pt,
  bottom=6pt,
  breakable
}
\definecolor{linkblue}{RGB}{0,166,214}
\acmConference{}
\begin{document}

\title{CiteAudit: You Cited It, But Did You Read It? \\ A Benchmark for Verifying Scientific References in the LLM Era}  

\author{\vspace{2pt} Kaiwen Shi$^{\text{1}}$\enskip Weixiang Sun$^{\text{1}}$\enskip Zheyuan Zhang$^{\text{1}}$\enskip Lichao Sun$^{\text{2}}$\enskip Nitesh V. Chawla$^{\text{1}}$\enskip Yanfang Ye$^{\text{1}}$\vspace{5pt}}\authornote{Corresponding author: yye7@nd.edu}
\affiliation{\vspace{1mm} $^{\text{1}}$University of Notre Dame \enskip $^{\text{2}}$Lehigh University \country{~}\vspace{5pt}}


\begin{abstract}
Scientific research is the fundamental driver of human societal progress, and proper citation is vital for attribution and research integrity. 
However, the rise of large language models (LLMs) has introduced a new integrity risk: fabricated references that appear plausible but correspond to no real publications. Recent analyses have uncovered such hallucinated citations even in submissions and accepted papers at major machine learning venues, underscoring growing vulnerabilities in peer-review workflows and raising concerns about the credibility of scholarly discourse. At the same time, rapidly expanding reference lists render manual verification infeasible, while existing automated tools remain fragile to the noise and formatting variability of real-world citation data and lack standardized, transparent evaluation.

This paper addresses these challenges by introducing the first comprehensive benchmark and detection framework for hallucinated citations in scientific writing. We design a multi-agent verification pipeline that decomposes citation checking into citation metadata extraction, memory lookup, web-based evidence retrieval, scholar search, and final judgment. This pipeline assesses whether a cited reference corresponds to a valid scholarly record and whether its core bibliographic fields are consistent with authoritative evidence. We further construct a large-scale, human-validated dataset spanning diverse domains, citation formats, and hallucination types, with unified evaluation protocols for citation existence and metadata consistency. Experiments with state-of-the-art LLMs and existing citation verification tools reveal substantial citation-related errors and show that our framework achieves stronger overall verification performance than commercial and open-source baselines. Our work provides systematic infrastructure for auditing citations at scale in the LLM era, helping researchers, reviewers, and publishers strengthen the trustworthiness of scientific references. Our code is available \href{https://github.com/shiiiikw/CiteAudit}{here}.

\end{abstract}

\maketitle

\section{Introduction}

\begin{figure}[t]
    \centering
    \includegraphics[width=\linewidth]{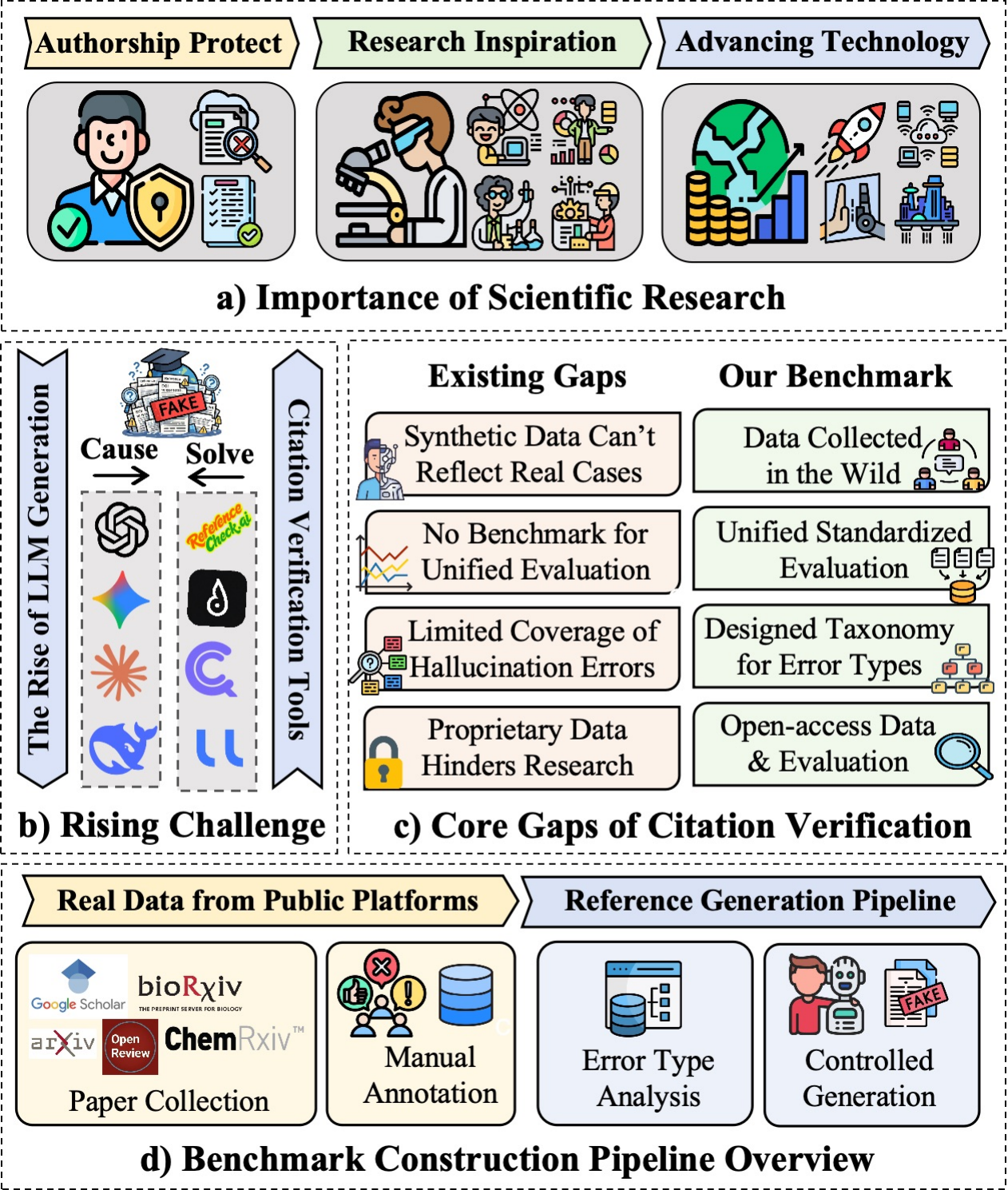}
    \caption{Motivation overview of our citation hallucination benchmark: existing gaps in closed-source citation checking tools and the unified, reliable evaluation framework enabled by our benchmark.}
    \label{fig:motivation}
    \vspace{-15pt}
\end{figure}

Scientific research constitutes the most critical engine of human progress, and the protection of authorship represents a fundamental respect for scholarly contributions as well as a vital source of intellectual inspiration. When citations are missing, inaccurate, or fabricated, the resulting break in the evidence chain can obscure logical dependencies, weaken argumentation, and jeopardize academic integrity at the level of both individual papers and the broader literature \cite{waltman2016review, chelli2024hallucination, sarol2024assessing}. However, the rapid adoption of large language models (LLMs) \cite{ye2025llms4all} and other generative systems \cite{yuan2024mora} has introduced a qualitatively new risk: the automatic creation of bibliographic entries that have no counterpart in the scholarly record. In contrast to conventional citation mistakes, such as incomplete metadata or minor typographical errors, hallucinated citations are entirely fabricated references that nevertheless resemble legitimate academic works. Recent investigations have documented such cases across prominent machine learning conferences, revealing hallucinated references in multiple submissions ~\cite{gptzero_iclr2026}. Independent reports further indicate that similar problems have surfaced even in accepted papers at leading forums such as NeurIPS and ACL~\cite{register2026neurips_hallucinations, sakoi2026hallucitation}. These incidents compromise multiple layers of the research process: they impede reviewers' ability to assess evidence, expose co-authors to inadvertent integrity violations, and weaken the reliability of the publication ecosystem as a whole, with downstream implications for reproducibility and the credibility of scientific discourse.

The growing scale of scholarly publishing further complicates this problem: reference lists have expanded rapidly across disciplines~\cite{dai2021literary}, making thorough manual verification unrealistic for reviewers, editors, and co-authors. This has motivated the development of automated citation-auditing tools \cite{gptzero,citely,swanref}. However, \textbf{prior works generally demonstrate two major gaps}: 1) Citation verification inevitably relies on retrieving information from external sources, yet the inherent noise and formatting variability of real-world references exposes a core weakness of existing systems, which frequently misfire when citations deviate from clean, canonical forms. 2) Most of these systems are proprietary, where they neither released their mechanism nor, more importantly, a large-scale, standardized and reproducible benchmark for hallucinated citation detection.

To bridge these two gaps, we introduce, to the best of our knowledge, the first comprehensive benchmark and detection framework for hallucinated citations in scientific manuscripts. Our contributions lay the groundwork for scalable tools that can support reviewers, editors, and automated review systems in upholding scholarly rigor in the LLM era. Specifically, we present a multi-agent framework and accompanying benchmark for verifying the existence and metadata consistency of scientific references. \textbf{To cope with the first challenge}, our system decomposes citation verification into cooperative roles: an Extractor parses citation strings into structured metadata, a Memory Agent reuses previously verified records, a Web Search Agent retrieves external evidence, a Scholar Agent queries authoritative scholarly sources, and a Judge Agent produces the final real-or-fake decision. This pipeline enables fine-grained assessment of whether a cited reference corresponds to a valid scholarly record and whether its title, authors, venue, year, and identifiers are consistent with external evidence. \textbf{To address the second challenge}, we construct a large-scale benchmark spanning diverse domains, citation formats, and hallucination types, with human-validated labels for both generated and real-world citation errors. The generated portion is built through controlled perturbations of verified references, while the real-world portion is collected from scholarly manuscripts and manually verified by the author team. We introduce unified evaluation protocols and metrics that measure citation existence, metadata consistency, classification performance, runtime, and cost across models. Experiments over leading LLMs and citation-verification tools reveal substantial rates of citation errors, including fabricated titles, incorrect authorship, venue mismatches, and invalid identifiers. Our analysis shows that the proposed multi-agent verification framework achieves stronger overall detection performance than commercial and open-source baselines. This work provides systematic infrastructure to audit citations at scale in the LLM era, offering a practical tool for researchers, reviewers, and publishers to assess and improve the trustworthiness of scientific references. Our contributions can be summarized as follows:

\begin{itemize}[left=0pt]
    \item \textbf{Benchmark.} We release the first large-scale, standardized benchmark for hallucinated citation detection, covering diverse domains and citation types with human-validated labels and unified evaluation protocols.

    \item \textbf{Framework.} We introduce a multi-agent verification pipeline that separates claim extraction, retrieval, matching, reasoning, and judgment, enabling robust citation checking under noisy and heterogeneous real-world formats.

    \item \textbf{Findings.} Through extensive experiments on state-of-the-art LLMs, we uncover pervasive citation errors and show that our framework yields stronger accuracy and interpretability than existing baselines.
\end{itemize}

\section{Related Works}

\subsection{Reference Hallucination Detection}
Reliably identifying AI-generated hallucinated content has become an increasing concern for both academia and industry \cite{huang2025survey,anh2025survey}. Among them, one of the most concerning risks lies in hallucinations in academic writing, where LLMs generate non-existent references \cite{chelli2024hallucination}. Such errors undermine scholarly trust and threaten the integrity of scientific communication \cite{tonmoy2024comprehensive,anh2025survey}. To verify the authenticity of academic references, some works \cite{gptzero,citely,swanref} have emerged to audit references by parsing citation strings and matching them against external bibliographic databases. Yet retrieval-based citation checking pipelines remain brittle to noise and variability inherent in real-world references, thus limiting their performance. To address this, more recent systems \cite{citecheck, refcheck_ai} adopt fuzzy matching strategies that compare citation fields against retrieved records using token-level similarity rather than exact string matching, enabling detection of mutated or incomplete references. However, these approaches still fundamentally reduce verification to field-level similarity matching, thus 
often failing under subtle or incomplete reference perturbations. More recent research begins to combine LLM-based reasoning models with retrieval for citation verification \cite{janse2025ai}, but these early efforts rely on overly limited and homogeneous external database sources, which can lead to false positive errors in practice. Moreover, their applicability can be further challenged in realistic settings where references must be extracted and verified from complex multimodal scholarly documents.

\subsection{Web Search Agent and Fact Checking}
LLM-based agents have recently demonstrated strong performance on complex, long-horizon tasks by moving beyond pure text generation toward actionable decision-making pipelines that interleave reasoning with interactions in external environments \cite{zhang2025agentrouter, shi2025ng, zhang2025mapro, hu2025hiagent,ferrag2025llm}. A key advantage of agentic systems over standalone foundation models is tool use—the ability to invoke external modules to acquire up-to-date evidence, execute operations, and reduce reliance on parametric memory. A representative and widely adopted form of tool use is web search \cite{nakano2021webgpt, ma2025autodata, yao2022react}, which enables agents to ground answers in retrieved evidence and thereby mitigate hallucinations \cite{zhang2025mopi, zhang2025ngqa}. Early works have further applied web search agents to fact-checking settings, demonstrating their effectiveness in evidence-based misinformation detection \cite{tian2024web}. In the context of citation verification, these advances in web search agent architectures motivate moving beyond approaches that rely solely on limited bibliographic APIs: by harnessing broader web search agents, systems can access a more comprehensive and diverse set of sources, thereby mitigating the coverage limitations inherent in API-based citation checks and improving the robustness of citation hallucination detection.
\section{Benchmark}



While a growing set of citation-verification systems has been developed to detect hallucinated citations in academic writing, many of these systems remain closed-source and proprietary, rendering their verification mechanisms opaque and their empirical performance irreproducible \cite{rahman2026hallucination}. This lack of transparency hinders systematic advancement: without an open, standardized benchmark, it’s infeasible to fairly compare methods or establish consistent evaluation protocols, thereby constraining the field’s scientific advancement, highlighting the need for a controlled, comprehensive, and reproducible benchmark for citation verification.

To address this gap, we introduced CiteAudit, a benchmark grounded in hallucinated citations reported on OpenReview. Specifically, we manually screen a large collection of papers, analyze the error patterns of citation hallucinations that naturally occur in real-world scholarly writing, and accordingly propose a structured taxonomy of fake citation types, with the detailed description in Appendix~\ref{sec:taxonomy}. Building on this foundation, our benchmark integrates both ecologically observed citation errors from the academic literature and controlled hallucinated references generated through principled perturbations. Benchmark statistics are shown in Table~\ref{tab:benchmark_stats}.

\begin{table}[t]
\centering
\caption{Statistical overview of our citation hallucination benchmark, including both the generated benchmark and the real-world test set.}\label{tab:benchmark_stats}
\renewcommand{\arraystretch}{1.2}
\setlength{\tabcolsep}{6pt}

\resizebox{\linewidth}{!}{%
\begin{tabular}{lccc}
\toprule
\textbf{Subset} & \textbf{Real} & \textbf{Fake}  & \textbf{Data Source} \\
\midrule
\begin{tabular}[c]{@{}l@{}}Generated\\Test Set\end{tabular}
& 3,586 & 2,500 
& \begin{tabular}[c]{@{}c@{}}GPT, Gemini, Claude Sonnet,\\ Qwen, Llama, etc \end{tabular}\\
\midrule
\begin{tabular}[c]{@{}l@{}}Real-World\\Test Set\end{tabular}
& 2,889 & 467 
& \begin{tabular}[c]{@{}c@{}}Google Scholar, OpenReview, \\ ArXiv, BioRxiv, etc \end{tabular}\\ 
\midrule
\textbf{Overall} 
& 6,475 & 2,967 
&  \\
\bottomrule
\end{tabular}
}
\end{table}

\subsection{Real World Data Collection}
We begin benchmark construction by collecting real-world citation entries from authentic scholarly manuscripts. 
Specifically, inspired by recent hallucination incidents reported at ICLR and NeurIPS, we collect citation instances from academic papers and scholarly records indexed in OpenReview, Google Scholar, arXiv, bioRxiv, and other public preprint or bibliographic platforms. 
From these sources, we systematically sample a large set of papers to obtain representative citation entries. We then cross-check their title, author list, venue, year, and other bibliographic metadata against authoritative scholarly records, and label each entry as verifiably correct only when all key fields consistently match; otherwise, we mark the reference as containing genuine errors or hallucinated components at the field level.

This real-world citation benchmark provides a high-quality gold reference set that reflects naturally occurring citation mistakes, including incorrect author attributions, venue mismatches, and nonexistent references. However, this process also highlights a key limitation: manual citation verification is highly labor-intensive and difficult to scale the dataset to a sufficiently large size. 

\subsection{Human-Synthesized Data Generation}
To overcome this scalability gap, we introduce the second component of our benchmark construction: a systematic framework for generating large-scale, controlled hallucinated citations. 

\begin{figure}[htbp]
    \centering
    \resizebox{\linewidth}{!}{\includegraphics{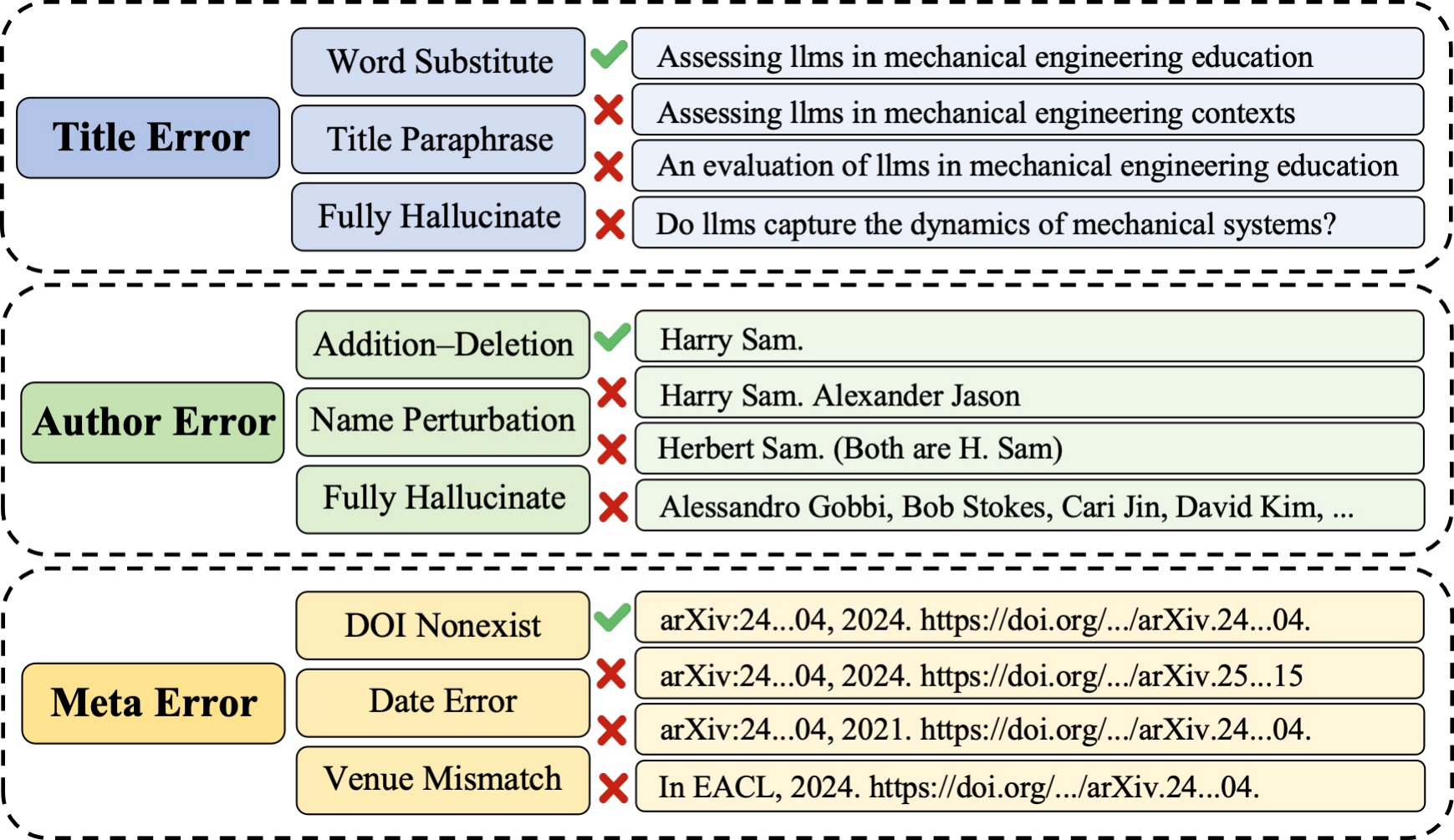}}
    \caption{Taxonomy of citation hallucination types.}
    \label{fig:taxonomy}
\end{figure}

\noindent\textbf{Real Citation Collection.} We first extract official BibTeX entries from publicly available open-access bibliographic repositories, spanning a broad spectrum of research areas and publication venues. These verified references serve as ground-truth citation records. Next, we generate hallucinated citations through targeted edits, guided by a principled taxonomy of citation hallucination types as shown in Figure~\ref{fig:taxonomy}. 

\noindent\textbf{Title Errors Generation.} Title hallucinations correspond to cases where the cited paper's title is incorrect or fabricated, while the remaining bibliographic fields remain unchanged. This error type is especially challenging because citation checkers often rely on fuzzy title matching, and minor title variations may appear plausible to both humans and automated systems. In our benchmark, we generate title errors through three complementary strategies. First, we perform keyword substitution, where core technical terms in the original title are replaced with semantically related alternatives, producing subtle but invalid variations. Second, we apply paraphrasing-based perturbations, where the title is rewritten into a fluent alternative expression that preserves topical coherence but does not correspond to any existing publication. Third, we introduce topic-conditioned fabrication, where a generative language model synthesizes a realistic-looking title within the same research area, ensuring that the hallucinated reference remains highly plausible despite being nonexistent. Together, these three strategies form an increasing spectrum of title hallucination severity, capturing hallucinations where generated citations appear semantically appropriate but fail existence verification.

\noindent\textbf{Author Error Generation.} Author hallucinations occur when the author list of a citation is partially or fully incorrect. Unlike superficial formatting errors, such mistakes directly distort scholarly attribution, potentially crediting nonexistent researchers or misassigning contributions, thereby undermining academic integrity and the reliability of citation-based verification. We construct author errors via four perturbation operations. First, we simulate authorship perturbations through redundant author additions that insert nonexistent names into correct author lists and through author deletions that remove valid authors, thereby producing incomplete attribution. Then, we apply name-level perturbations to the existing author list, introducing realistic identity inconsistencies through edits to author name strings, such as swapping given and family names, while preserving the overall author set structure. Finally, we construct fully synthetic author lists, where the entire set of authors is fabricated rather than derived from the original reference. These author perturbations enable systematic evaluation of citation checkers' sensitivity to identity-level inconsistencies.

\noindent\textbf{Metadata Error Generation.} Metadata hallucinations involve incorrect bibliographic fields beyond title and authors, such as venue names, publication years, or persistent identifiers. These errors reflect cases where a model recalls a paper approximately but misattributes its publication context. We generate metadata errors by perturbing key BibTeX fields. Venue errors are constructed by replacing the true conference or journal name with a related but incorrect outlet. Year errors are introduced by shifting publication dates to plausible but invalid alternatives. In addition, we generate DOI and identifier hallucinations, where DOI strings are fabricated or mismatched, simulating errors that cannot be detected through surface-level text similarity alone. Metadata errors therefore capture hallucination patterns where citations appear structurally complete but fail bibliographic consistency checks.

Our generated benchmark exhibits error characteristics that are highly consistent with those observed in real-world citation hallucinations. 
As shown in Table~\ref{tab:gptzero_chi}, we compare GPTZero's fake-citation detection behavior on hallucinated citations between the generated benchmark and the real-world set. 
A chi-square test shows no significant difference ($\chi^2 = 5.6 \times 10^{-5}$, $p = 0.994$), indicating highly consistent behavior across the two settings and supporting the fidelity of our generation framework in simulating real-world citation hallucination patterns. 
Additional statistics and breakdowns of the generated dataset are provided in Appendix~\ref{sec:dataset}.

\begin{figure*}
    \centering
    \includegraphics[width=1\linewidth]{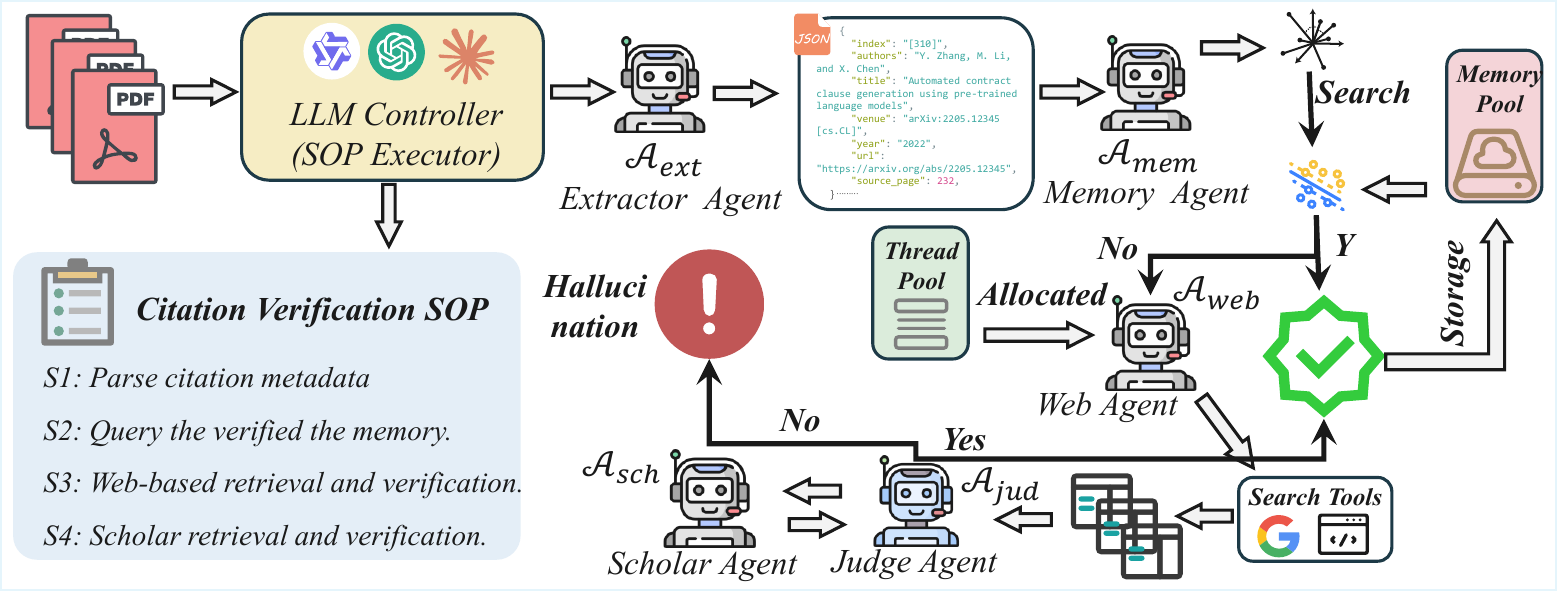}
    \caption{Overview of CiteAudit, a SOP-driven multi-agent citation verification framework.}
    \label{fig:placeholder}
\end{figure*}

\begin{table}[t]
\centering
\renewcommand{\arraystretch}{1.15}
\setlength{\tabcolsep}{6pt}
\caption{Chi-square comparison of GPTZero fake-citation detection between generated and real-world datasets.}
\resizebox{\linewidth}{!}{%
\begin{tabular}{lccc}
\toprule
\textbf{Dataset} & \textbf{Pred Fake} & \textbf{Pred Real} & \textbf{Total} \\
\midrule
Generated Test Set Fake Citations & 1809 & 691 & 2500 \\
Real-World Test Set Fake Citations & 338 & 129 & 467 \\
\midrule
\multicolumn{4}{c}{$\chi^2 = 5.6 \times 10^{-5},\; p = 0.994\; (df = 1)$} \\
\bottomrule
\end{tabular}
}
\label{tab:gptzero_chi}
\end{table}






\subsection{Data Annotation}
To ensure benchmark reliability, both the real-world citation entries collected in the data collection stage and the hallucinated citations generated through perturbation are subjected to a rigorous multi-step annotation and verification pipeline. We adopt a screening process that combines automated evidence retrieval with careful human validation. First, a web-search-based model is employed to retrieve corresponding online evidence for each citation, linking the bibliographic fields to relevant publication pages or authoritative scholarly records. This retrieval step provides scalable support for large volumes of citations by surfacing candidate sources for verification. Subsequently, the author team manually inspects the retrieved evidence and conducts detailed cross-checking to confirm citation authenticity and resolve potential mismatches. Through this human-in-the-loop verification, we ensure that each benchmark entry is assigned an accurate and high-confidence label, indicating whether it corresponds to a real reference or a hallucinated/erroneous citation. Through this combination of web-grounded retrieval and rigorous author-led validation, our benchmark offers a systematic and reliable resource for evaluating citation verification systems under both naturally occurring citation errors and diagnostically controlled hallucination scenarios. Detailed human annotation guidelines and verification procedures are provided in Appendix~\ref{sec:annotation}.

\section{Methodology}

In this section, we formalize the detection of hallucinated citations as a multi-stage evidence verification problem. We introduce a decentralized multi-agent framework coordinated via a hierarchical Standardized Operating Procedure (SOP), designed to systematically audit scholarly references for both existence and metadata integrity.

\subsection{Problem Formulation}
Let $\mathcal{D}$ be a scientific document containing a set of citation strings $\mathcal{R} = \{r_1, r_2, \dots, r_n\}$. For each citation $r_i$, our objective is to determine a binary verdict $v_i \in \{Fake, Real\}$. We represent each citation as a structured metadata tuple $M_i = \{m_T, m_A, m_U, m_V, m_Y\}$, where fields denote title, authors, URL or identifier, publication venue, and publication year, respectively. We define the verification function $\mathcal{F}: M_i \rightarrow \{0, 1\}$ based on a \textbf{Metadata Consistency Criterion} $S_c$:
\begin{equation}
S_c = \prod_{k \in \{T, A, U, V, Y\}} \mathbb{I}\big(\text{Consistent}(m_k, \hat{m}_k)\big)
\end{equation}
where $\hat{m}_k$ represents the metadata retrieved from authoritative databases, and $\mathbb{I}(\cdot)$ is the indicator function that outputs 1 if and only if the extracted field $m_k$ is consistent with the retrieved evidence under predefined matching rules, and 0 otherwise. These rules use strict title matching for academic publications, lenient author matching under standard name normalization, and reasonable venue and year tolerance for preprint-to-publication variants. A citation is classified as $Fake$ if no corresponding entry exists in the global scholarly graph $\mathcal{G}_{scholar}$ or if $S_c = 0$.

\subsection{Collaborative Multi-Agent Pipeline}
Our framework instantiates five specialized agents, each governed by a restricted action space and a specific role in the verification pipeline.

\noindent\textbf{Extractor Agent ($\mathcal{A}_{ext}$):} The verification pipeline is initiated by $\mathcal{A}_{ext}$, which acts as a vision-integrated structural parser. Instead of traditional linguistic analysis, this agent orchestrates high-precision OCR tools such as Nougat and PyMuPDF to ingest raw text and visual coordinates from the PDF manuscript. The LLM then performs a schema-constrained transformation to map these unformatted strings into an immutable metadata set $M_i = \{m_T, m_A, m_U, m_V\}$. This ensures that the original citation as presented by the author is preserved with minimal semantic distortion, providing the raw substrate for downstream metadata consistency auditing.

\noindent\textbf{Dual-End Memory Agent ($\mathcal{A}_{mem}$):} To minimize redundant computation, $\mathcal{A}_{mem}$ executes a semantic lookup across a dual-end knowledge base $\mathcal{K}$. We formalize the retrieval as a vector similarity function. Let $Enc(\cdot)$ be the embedding model, the agent computes the confidence score $s_{mem}$:
\begin{equation}
s_{mem}(M_i) = \max_{k \in \mathcal{K}} \left( \frac{Enc(M_i) \cdot Enc(k)}{||Enc(M_i)|| \cdot ||Enc(k)||} \right)
\end{equation}
If $s_{mem} > \tau$ where $\tau=0.92$, the citation is immediately verified via the fast-path, bypassing external retrieval.

\noindent\textbf{Web Search Agent ($\mathcal{A}_{web}$):} In scenarios where internal memory lookup results in a cache miss, $\mathcal{A}_{web}$ is triggered to conduct external validation. This agent interfaces with the Google Search API to identify relevant online evidence for the cited reference. Rather than relying solely on search snippets, $\mathcal{A}_{web}$ retrieves content from the top search results and extracts candidate evidence from scholarly or bibliographic sources. By ingesting textual data from author homepages, institutional repositories, publisher pages, and preprint platforms, the agent ensures that the subsequent judgment is grounded in retrieved evidence rather than superficial link descriptions.

\noindent\textbf{Judge Agent ($\mathcal{A}_{jud}$):} As the central decision engine, $\mathcal{A}_{jud}$ evaluates the alignment between the extracted metadata $M_i$ and the retrieved evidence set $\mathcal{E}$. We define the verification function $\mathcal{F}_{judge}$ as:
\begin{equation}
\mathcal{F}_{judge}(M_i, \mathcal{E}) = \prod_{f \in \{T, A, U, V\}} \mathbb{I}\big(\text{Consistent}(M_i^f, \mathcal{E})\big)
\end{equation}
where $\mathbb{I}(\cdot)$ is the indicator function. The agent acts as a gatekeeper: it returns $Real$ only when the retrieved evidence satisfies the predefined title, author, venue, and identifier consistency rules from $\mathcal{A}_{web}$ or $\mathcal{A}_{sch}$.

\noindent\textbf{Scholar Agent ($\mathcal{A}_{sch}$):} Acting as the \textit{Veracity Benchmark}, this agent is invoked for high-stakes validation when preliminary web evidence is insufficient. It executes targeted, low-frequency crawling of authoritative repositories such as Google Scholar, publisher pages, DOI records, and preprint repositories to retrieve the canonical record $\hat{M}_i$. 

\subsection{Collaborative Multi-Agent Pipeline and Planning Model}
In this section, we describe the system's execution kernel, which is orchestrated by an LLM Controller acting as the central SOP Executor. The coordination logic follows a rigorous task-allocation model governed by a Planning Model, ensuring high throughput via the Multi-Thread.

\noindent\textbf{Planning Model:}
The orchestrator is the structural backbone of the framework. Given a PDF manuscript, the Planning Model decomposes the verification job into a sequential and parallelizable graph based on the predefined Citation Verification SOP (S1-S4). It manages the state transitions of each citation task, ensuring that resources are optimally allocated from the Thread Pool. 

\noindent\textbf{Formalization of SOP Execution.}
The coordination logic described in Stages 1-4 can be formalized as a hierarchical cascade function $\Phi(r_i)$. The Planning Model routes the verification task sequentially to optimize the trade-off between cost and accuracy:

\begin{table*}[t]
\centering
\caption{
Results on the generated test set. Runtime efficiency, API pricing, and verification performance of citation verification models.
Prices are reported per one million tokens based on official or primary provider APIs when available.
As GPTZero pricing is defined per word in subscription tiers, a coarse token-level estimate is used for comparison.
}
\renewcommand{\arraystretch}{1.15}
\setlength{\tabcolsep}{4pt}

\begin{tabular}{l c cc cccc cccc}
\toprule
\textbf{Model}
& \textbf{Time}
& \multicolumn{2}{c}{\textbf{Price (\$/1M tok)}}
& \multicolumn{4}{c}{\textbf{Confusion Matrix}}
& \multicolumn{4}{c}{\textbf{Metrics}} \\
\cmidrule(lr){3-4} \cmidrule(lr){5-8} \cmidrule(lr){9-12}
& \textbf{/10 refs}
& \textbf{In} & \textbf{Out}
& \textbf{TP} & \textbf{FN} & \textbf{FP} & \textbf{TN}
& \textbf{Acc} & \textbf{Prec} & \textbf{Rec} & \textbf{F1} \\
\midrule

Mixtral-8x7B-Instruct
& 2.3 & 0.60 & 0.60
& 1675 & 825 & 940 & 2646
& 0.710 & 0.641 & 0.670 & 0.655 \\

Llama-3.3-70B-Instruct
& 4.9 & 0.88 & 0.88
& 1088 & 1412 & 381 & 3205
& 0.705 & 0.741 & 0.435 & 0.548 \\

Qwen3-Next-80B-A3B
& 3.5 & 0.15 & 1.50
& 1265 & 1235 & 1370 & 2216
& 0.572 & 0.480 & 0.506 & 0.492 \\

Gemini-3-Pro
& 36.6 & 2.00 & 12.00
& 1879 & 621 & 511 & 3075
& 0.814 & 0.786 & 0.752 & 0.769 \\

GPT-5.2
& 47.1 & 1.75 & 14.00
& 2284 & 216 & \textbf{0} & \textbf{3586}
& 0.965 & \textbf{1.000} & 0.914 & 0.955 \\

GPTZero
& 26.3 & 70.00 & 0.00
& 1809 & 691 & 623 & 2963
& 0.784 & 0.744 & 0.724 & 0.734 \\

Claude-Sonnet-4.5
& 11.3 & 3.00 & 15.00
& \textbf{2475} & \textbf{25} & 3364 & 222
& 0.443 & 0.424 & \textbf{0.990} & 0.594 \\

\midrule
\textbf{Our Model}
& 11.2 & 0.50 & 3.00
& 2428 & 72 & 136 & 3450
& \textbf{0.966} & 0.947 & 0.971 & \textbf{0.959} \\

\bottomrule
\end{tabular}
\label{tab:generated_performance}
\end{table*}

\begin{equation}
\Phi(r_i) = 
\begin{cases} 
\text{Verified} & \text{if } \mathcal{A}_{mem}(r_i) > \tau \quad \text{(Stage 2)} \\
\text{Verified} & \text{if } \mathcal{A}_{jud}(r_i, \mathcal{A}_{web}) = 1 \quad \text{(Stage 3)} \\
\mathcal{A}_{jud}(r_i, \mathcal{A}_{sch}) & \text{otherwise} \quad \text{(Stage 4)}
\end{cases}
\end{equation}

This formulation ensures that the computationally expensive \textit{Scholar Agent} ($\mathcal{A}_{sch}$) is only invoked as a fallback for unresolved or ambiguous citations, adhering to the principle of minimal resource consumption.

\noindent\textbf{Stage 1: Citation Metadata Extraction ($\mathcal{A}_{ext}$):}
The process is initiated by $\mathcal{A}_{ext}$, which maps the visual and textual data from the PDF into a structured JSON schema. This transformation converts implicit citations into explicit, verifiable metadata ${m_T, m_A, m_U, m_V}$, providing the raw data substrate for the entire pipeline.

\noindent\textbf{Stage 2: Verified Memory Querying ($\mathcal{A}_{mem}$):}
To optimize latency, the controller routes the metadata to $\mathcal{A}_{mem}$. This agent performs a high-speed lookup in the Memory Pool. If the reference has been previously audited and marked as \textit{Verified} ($Y$), the task concludes immediately. If not found ($N$), the planning model allocates the task to the next tier of the thread pool.

\noindent\textbf{Stage 3: Web-based Content Retrieval \& Consistency Audit:}
Uncached citations are processed by the Web Search Agent and the Judge Agent. The orchestrator allocates parallel threads to retrieve high-relevance evidence from the live web. The Judge then performs the predefined metadata consistency check. A successful match ($Y$) updates the Memory Pool for future reuse, while a mismatch or inconclusive result ($N$) triggers an escalation to the final verification tier.

\noindent\textbf{Stage 4: Scholar Retrieval \& Final Verification:}
Unresolved citations are handled by the Scholar Agent. It utilizes low-frequency, high-precision crawling to fetch canonical scholarly records. The Judge Agent then performs a second-pass rule-based metadata consistency check. If this definitive check fails ($N$), the citation is flagged as \textit{Fake} with an associated provenance report; otherwise, it is successfully marked as \textit{Verified} and stored.
\section{Experiments}

\begin{table*}[t]
\centering
\caption{
Results on the real-world test set. The results show strong consistency with the generated test set, supporting the external validity of the generated benchmark.
}
\label{tab:realworld_performance}

\renewcommand{\arraystretch}{1.15}
\setlength{\tabcolsep}{4pt}

\begin{tabular}{l c cc cccc cccc}
\toprule
\textbf{Model}
& \textbf{Time}
& \multicolumn{2}{c}{\textbf{Price (\$/1M tok)}}
& \multicolumn{4}{c}{\textbf{Confusion Matrix}}
& \multicolumn{4}{c}{\textbf{Metrics}} \\
\cmidrule(lr){3-4} \cmidrule(lr){5-8} \cmidrule(lr){9-12}
& \textbf{/10 refs}
& \textbf{In} & \textbf{Out}
& \textbf{TP} & \textbf{FN} & \textbf{FP} & \textbf{TN}
& \textbf{Acc} & \textbf{Prec} & \textbf{Rec} & \textbf{F1} \\
\midrule

Mixtral-8x7B-Instruct
& 2.3 & 0.60 & 0.60
& 95  & 372 & 757  & 2132
& 0.664 & 0.112 & 0.203 & 0.144 \\

Llama-3.3-70B-Instruct
& 4.8 & 0.88 & 0.88
& 83  & 384 & 306  & 2583
& 0.794 & 0.213 & 0.178 & 0.194 \\

Qwen3-Next-80B-A3B
& 3.7 & 0.15 & 1.50
& 234 & 233 & 1104 & 1785
& 0.602 & 0.175 & 0.501 & 0.259 \\

Gemini-3-Pro
& 38.1 & 2.00 & 12.00
& 351 & 116 & 412  & 2477
& 0.843 & 0.460 & 0.752 & 0.571 \\

GPT-5.2
& 48.8 & 1.75 & 14.00
& 366 & 101 & 1379 & 1510
& 0.559 & 0.210 & 0.784 & 0.331 \\

GPTZero
& 26.2 & 70.00 & 0.00
& 338 & 129 & 1358 & 1531
& 0.557 & 0.199 & 0.724 & 0.313 \\

Claude-Sonnet-4.5
& 13.3 & 3.00 & 15.00
& 349 & 118 & 756  & 2133
& 0.740 & 0.316 & 0.747 & 0.444 \\

\midrule
\textbf{Our Model}
& 11.2 & 0.50 & 3.00
& \textbf{444} & \textbf{23} & \textbf{149} & \textbf{2740}
& \textbf{0.949} & \textbf{0.749} & \textbf{0.951} & \textbf{0.838} \\

\bottomrule
\end{tabular}
\end{table*}
\subsection{Experimental Setup}

In this section, we detail the implementation of our multi-agent framework and the hardware/software configurations used for the citation verification task.

\noindent\textbf{Agent Implementation and Model Selection:}
Our framework combines multimodal citation extraction, external evidence retrieval, memory-based reuse, and model-based judgment. 
The extraction component is implemented with \textbf{Qwen3-VL-235B A22}, while the planning and final judgment components are powered by \textbf{Gemini 3 Flash}. 
The specific roles are instantiated as follows:

\begin{itemize}[left=0pt]
\item \textbf{Planning Model (Orchestrator):} Implemented using \textbf{Gemini 3 Flash}. It acts as the central executor of the SOP, managing task states and thread-pool allocation. It coordinates citation extraction, evidence retrieval, memory lookup, scholarly search, and final judgment.

\item \textbf{Extractor Agent ($\mathcal{A}_{ext}$):} Utilizing the multimodal capabilities of \textbf{Qwen3-VL-235B A22} \cite{bai2025qwen3vltechnicalreport}, this agent performs page-level OCR and structural parsing. It identifies citation strings and maps them into a predefined JSON metadata schema.

\item \textbf{Memory Agent ($\mathcal{A}_{mem}$):} Developed based on the \textbf{Mem0} \cite{mem0} framework. It maintains a persistent, evolving knowledge graph of previously audited citations, enabling long-term context retention and rapid fast-path verification.

\item \textbf{Web Search Agent ($\mathcal{A}_{web}$):} Integrated with the Google Search API for real-time evidence retrieval. It issues metadata-based search queries and collects candidate evidence from web-accessible scholarly and bibliographic sources.

\item \textbf{Judge Agent ($\mathcal{A}_{jud}$):} Powered by \textbf{Gemini 3 Flash}, this agent executes the Strict Consistency Criterion ($S_c$). It compares extracted citation metadata against retrieved external evidence and acts as the final arbiter for both preliminary and escalated verification stages.

\item \textbf{Scholar Agent ($\mathcal{A}_{sch}$):} A specialized high-precision crawler \cite{playwright_python} designed to interface with authoritative scholarly databases, such as Google Scholar, to fetch canonical ground-truth records ($\hat{M}_i$).
\end{itemize}

The detailed system prompts and standardized operating procedure (SOP) definitions for all agents are documented in Appendix~\ref{sec:prompts}.

\noindent\textbf{Infrastructure and Hyperparameters:}
The extraction pipeline is executed on a high-performance compute cluster equipped with NVIDIA B200 GPUs, while the planning and judgment stages use Gemini 3 Flash through API inference. We employ a \textbf{Multi-thread Pool (Size=4)} to facilitate simultaneous citation auditing across multiple documents. For the visual-textual extraction and final judgment stages, decoding is configured with a temperature of $0.0$ to ensure deterministic and reproducible parsing and verification. The vector database for $\mathcal{A}_{mem}$ utilizes cosine similarity with a threshold of $0.92$ to identify high-affinity citation matches.

\noindent\textbf{Evaluation Metrics.}  
To evaluate the performance of our citation verification system, we report both classification-based and efficiency-oriented metrics.

We define four types of prediction outcomes. \textbf{True Positives} refer to hallucinated citations that are correctly flagged as fake. \textbf{False Negatives} are hallucinated citations that are mistakenly predicted as real. \textbf{False Positives} denote real citations that are incorrectly flagged as fake, and \textbf{True Negatives} are real citations that are correctly identified as real.

Based on these outcomes, we compute standard metrics. \textbf{Accuracy} reflects the overall correctness of the predictions across all citations. \textbf{Precision} measures how often the citations flagged as hallucinations are actually hallucinated, indicating the system’s ability to avoid false alarms. \textbf{Recall} quantifies how many of the actual hallucinated citations are successfully identified, capturing detection completeness. The \textbf{F1 score} balances precision and recall, offering a single measure of effectiveness in hallucination detection.

In addition to classification performance, we also assess system efficiency by measuring the average \textbf{runtime} required to verify a batch of 10 citations, along with the associated input/output costs. These metrics reflect the practical feasibility of large-scale deployment and help benchmark the trade-off between verification accuracy and computational cost.

\subsection{Evaluation on Generated Benchmark}

We evaluate all citation verification models on our generated benchmark, which consists of 3,586 real-world references and 2,500 hallucinated references produced through controlled perturbations spanning multiple error categories. In addition, we conduct evaluation on a real-world test set comprising 2,889 authentic references and 467 naturally occurring hallucinated citations collected from real scholarly sources.

Table~\ref{tab:generated_performance} presents the performance of different citation verification models on our generated benchmark. 
Most existing models exhibit a clear imbalance between detecting hallucinated citations and preserving real references. For example, GPT-5.2 achieves perfect precision and makes no false-positive errors on real citations, but it misses a larger number of hallucinated references than the highest-recall systems. Conversely, Claude-Sonnet-4.5 achieves the highest recall, but introduces a large number of false positives on genuine references. 
In contrast, our model achieves the best overall balance between hallucination detection and real-reference preservation. It obtains the highest accuracy and F1 score, while maintaining both high precision and high recall. This indicates that our approach avoids both overly permissive acceptance of fabricated references and overly aggressive rejection of genuine citations, leading to more reliable citation verification performance.

Table~\ref{tab:generated_performance} also reports runtime efficiency and API pricing across representative citation-verification models. Our framework does not achieve the lowest runtime or the lowest API price, but it offers a favorable trade-off between cost, latency, and verification quality. In particular, it achieves the strongest overall classification performance on the generated benchmark while using substantially cheaper inference than several proprietary high-end models. This suggests that the proposed agentic verification design can improve reliability without relying solely on the most expensive model calls.

\subsection{Evaluation on Real World Benchmark}

We additionally assess citation verification performance on a real-world collection composed of authentic references and naturally occurring hallucinated citations drawn from scholarly manuscripts.
Compared with the controlled benchmark setting, this evaluation reflects the ambiguity, noise, and incomplete metadata commonly encountered in practical academic writing scenarios.

The results in Table~\ref{tab:realworld_performance} reveal that existing approaches remain limited in balancing hallucination filtering with preservation of legitimate references. Methods that aggressively flag suspicious entries tend to over-reject genuine citations, whereas more permissive systems allow fabricated references to pass verification, indicating unresolved reliability challenges in real deployment conditions. In contrast, our framework achieves the strongest overall performance across all reported classification metrics.
Specifically, it delivers the highest accuracy, precision, recall, and F1 score, with the F1 score exceeding that of the second-best system by 0.267 absolute points. This margin indicates a substantial improvement in balanced verification capability. Our method detects hallucinated citations more reliably while maintaining faithful acceptance of genuine references, overcoming the reliability limitations observed in prior systems.

Moreover, the consistency of relative model behavior between this real-world evaluation and the controlled benchmark suggests that the perturbation-based construction pipeline captures important characteristics of citation errors occurring in practice, further supporting the empirical validity and diagnostic value of our benchmark design.

\subsection{Additional Experiment Analysis}
During experimentation, the suboptimal performance of advanced proprietary models appeared counterintuitive, motivating diagnostic evaluations via web-based LLM interfaces with observable reasoning behavior. We found that even when explicitly instructed to perform external retrieval, the systems do not reliably execute verifiable search procedures, and the provenance of implicitly retrieved evidence remains opaque. This black-box behavior makes retrieval neither enforceable nor transparently grounded, which is problematic for citation verification requiring explicit evidence tracing. This observation further underscores the necessity of specialized, auditable citation-verification tools that ground decisions in traceable external evidence, reinforcing the value of our benchmark and system as a principled alternative to reliance on closed-source general-purpose LLMs.

\subsection{Case Study}
\begin{figure}[t]
\centering
\includegraphics[width=\linewidth]{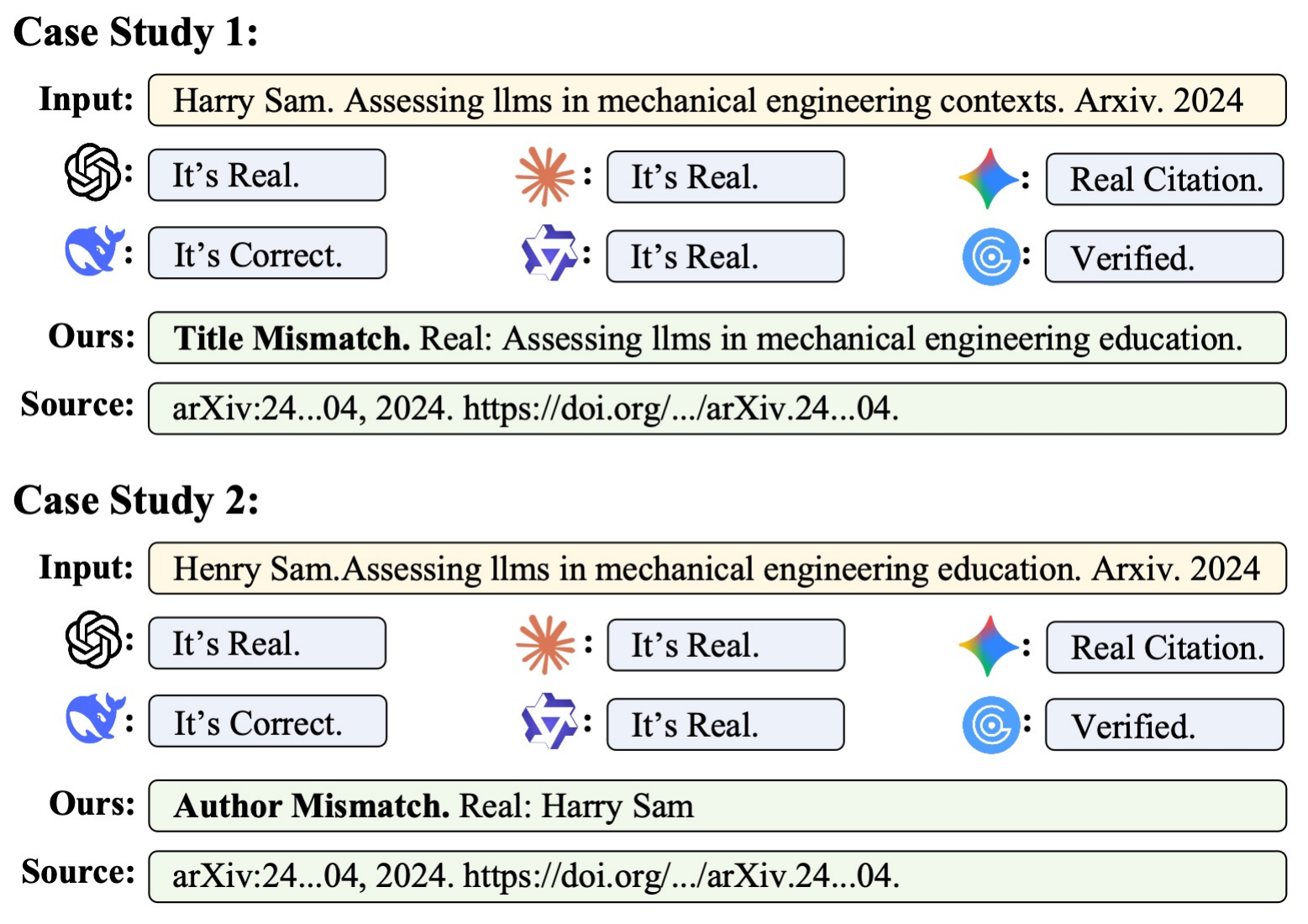}
\caption{Case studies of citation verification}
\label{fig:citation_case_study}

\end{figure}

We present representative case studies to qualitatively demonstrate the effectiveness and robustness of our citation verification framework. As illustrated in Figure~\ref{fig:citation_case_study}, the system is capable of accurately identifying whether a citation refers to a real scholarly work, while further diagnosing fine-grained inconsistency types when mismatches occur.

In \textbf{Case Study 1}, although the queried citation corresponds to a real arXiv paper and is correctly recognized as non-hallucinated by multiple baseline systems, the cited title exhibits a subtle semantic deviation from the ground-truth record. Our framework successfully detects this \emph{title mismatch}, retrieves the correct reference from arXiv, and explicitly reports the discrepancy, demonstrating sensitivity to partial but meaningful citation errors beyond binary real/fake classification. Similarly, in \textbf{Case Study 2}, the input citation again refers to an existing paper; however, the listed author name does not match the true authorship information. While existing tools label the citation as valid due to its overall plausibility, our system precisely identifies the \emph{author mismatch} and recovers the correct author metadata from the authoritative source.

Across both examples, the framework not only distinguishes real citations from hallucinated ones, but also performs structured verification of individual metadata fields, including title and author information. By retrieving the correct ground-truth source and reporting explicit mismatch categories, our approach enables reliable, interpretable, and practically useful citation authenticity verification, which is essential for real-world scholarly and clinical documentation workflows.

\section{Conclusion}

As large language models become deeply integrated into scientific writing and peer-review workflows, hallucinated citations pose a growing threat to research integrity. These references can appear plausible in title, authors, venue, or year while remaining nonexistent or inconsistent with real scholarly records, making them difficult to detect through surface inspection alone.

In this work, we introduce an open, standardized, and scalable benchmark for hallucinated citation detection, covering both controlled perturbations and naturally occurring real-world citation errors. We also propose a multi-agent framework that verifies citations through a structured SOP-driven process, combining planning, extraction, retrieval, memory-based reuse, scholarly search, and final judgment.

Our experiments show that existing commercial and open-source baselines often struggle to balance detecting hallucinated citations with preserving genuine references. In contrast, our framework achieves strong performance in both generated and real-world settings, demonstrating a more reliable balance across accuracy, precision, recall, and F1 score. Overall, our benchmark and system provide practical infrastructure for building more accountable citation verification tools in the LLM era.
\newpage
%
\bibliographystyle{ACM-Reference-Format}
\bibliography{sample-base}

\newpage

\appendix

\section{Citation Hallucination Taxonomy}
\label{sec:taxonomy}

To support systematic and reproducible evaluation of citation verification systems, we propose a structured taxonomy of citation hallucinations. The taxonomy categorizes hallucinations according to the type and severity of bibliographic inconsistencies. It is grounded in citation errors observed in real scholarly manuscripts and is designed to capture both subtle and severe hallucination patterns produced by large language models.

\subsection{Definition of Citation Hallucination}

We define a \textit{citation hallucination} as a bibliographic reference that appears plausible in form and content but does not correspond to a valid scholarly record. Formally, given a citation metadata tuple $M = \{m_T, m_A, m_U, m_V, m_Y\}$ representing title, authors, URL or persistent identifier, venue, and publication year, a citation is considered \textit{hallucinated} if no authoritative scholarly source can be found that satisfies all essential metadata fields.

Citation hallucinations differ from minor formatting inconsistencies or incomplete references. They involve semantic or factual mismatches that break the evidence chain between a claim and its cited source, thereby weakening scholarly attribution and verification.

\subsection{Title-Level Hallucinations}

Title-level hallucinations occur when the cited paper title is incorrect or fabricated, while other metadata fields remain partially or fully plausible. This error type is especially difficult because minor lexical variations or fluent paraphrases may appear legitimate to both human readers and automated systems.

We identify three representative subclasses:
\begin{itemize}[left=0pt]
    \item \textbf{Keyword Substitution:} Core technical terms in the original title are replaced with semantically related but incorrect alternatives.
    \item \textbf{Paraphrased Fabrication:} The title is fluently rephrased into a plausible variant that does not correspond to any existing publication.
    \item \textbf{Topic-Conditioned Synthesis:} A realistic-looking title is generated within the same research area without grounding in any real work.
\end{itemize}

A citation with any of these patterns is labeled as hallucinated if no exact or canonical title match can be verified in authoritative bibliographic databases.

\subsection{Author-Level Hallucinations}

Author-level hallucinations involve incorrect or fabricated authorship information. These errors directly distort scholarly attribution and can mislead citation-based evaluation and credit assignment.

We categorize author hallucinations into four types:
\begin{itemize}[left=0pt]
    \item \textbf{Author Addition:} Nonexistent or unrelated author names are inserted into an otherwise valid author list.
    \item \textbf{Author Deletion:} One or more legitimate authors are omitted from the citation.
    \item \textbf{Name Perturbation:} Author names are altered through spelling changes, reordered given and family names, or partial truncation.
    \item \textbf{Fully Fabricated Authorship:} The entire author list does not correspond to any real publication.
\end{itemize}

A citation is marked as hallucinated if the author list cannot be aligned with a verified scholarly record under standard name normalization rules.

\subsection{Metadata-Level Hallucinations}

Metadata-level hallucinations refer to inconsistencies in bibliographic fields beyond title and authors, including venue, publication year, DOI, arXiv ID, and other persistent identifiers.

We consider the following categories:
\begin{itemize}[left=0pt]
    \item \textbf{Venue Mismatch:} The cited venue does not match the actual publication outlet of the work.
    \item \textbf{Year Mismatch:} The publication year is incorrect beyond acceptable variation between preprint and final versions.
    \item \textbf{Identifier Fabrication:} DOI, arXiv ID, or other persistent identifiers are invalid or assigned to a different work.
\end{itemize}

These hallucinations are difficult to detect through surface-level similarity matching because the citation may appear structurally complete while still being bibliographically invalid.

\subsection{Compound and Cross-Field Hallucinations}

In practice, citation hallucinations often involve multiple simultaneous inconsistencies. We therefore include a compound category for citations that contain errors across two or more metadata fields, such as title and authors, or venue and identifier.

These compound hallucinations represent the most severe form of citation fabrication and are always labeled as hallucinated in our benchmark.

\subsection{Relation to Real-World Citation Errors}

Our taxonomy is informed by empirical analysis of citation errors observed in real conference and journal submissions. We emphasize that not all citation inaccuracies constitute hallucinations. Minor formatting variations, missing page numbers, and capitalization differences are not considered hallucinations as long as the core bibliographic identity remains verifiable.

This distinction ensures that the benchmark targets genuinely harmful hallucinations rather than benign citation noise, supporting fair and realistic evaluation of citation verification systems.

\section{Dataset Construction and Statistics}
\label{sec:dataset}

This appendix details the construction process and statistical composition of the generated citation hallucination dataset used in our benchmark.
The dataset is designed to support controlled and fine-grained evaluation of citation verification systems under diverse hallucination scenarios.

\subsection{Source of Real Citations}

We begin by collecting a pool of verified real citations from publicly available bibliographic repositories and open-access scholarly sources.
Each citation is extracted in structured BibTeX format and verified to ensure correctness across essential metadata fields, including title, authors, venue, publication year, DOI, arXiv ID, and other persistent identifiers.
These verified citations serve two purposes. First, they provide the real-reference portion of the generated benchmark. Second, they serve as seed references from which hallucinated variants are systematically generated.

\subsection{Hallucinated Citation Generation}

Building on the verified citation pool, we generate hallucinated references through controlled perturbations guided by the taxonomy described in Appendix~\ref{sec:taxonomy}.
Each hallucinated citation is derived from a real reference by modifying one or more metadata fields while preserving overall plausibility and valid citation structure.

The generated test set contains three primary hallucination categories:
\begin{itemize}[left=0pt]
    \item \textbf{Title Errors (1,000 instances):} The original title is replaced with a semantically plausible but invalid variant, while the remaining metadata fields are preserved.
    These errors are constructed through keyword substitution, paraphrase-based rewriting, and topic-conditioned synthesis.
    
    \item \textbf{Author Errors (1,000 instances):} The author list is perturbed through controlled operations, including author addition, author deletion, name-level perturbation, and full authorship fabrication, while the remaining metadata fields are preserved.
    
    \item \textbf{Metadata Errors (500 instances):} Bibliographic fields beyond title and authors are modified, including venue mismatches, publication year shifts, and fabricated or mismatched persistent identifiers.
\end{itemize}

Each hallucinated citation maintains valid BibTeX structure and formatting so that models cannot rely on trivial syntactic irregularities for detection.

\subsection{Dataset Composition and Balance}

The final generated test set contains 2,500 hallucinated citations and 3,586 verified real citations, resulting in 6,086 citation instances in total.
The hallucinated portion is distributed across title-level, author-level, and metadata-level errors as described above.
The real-reference portion is sampled from the verified citation pool and retained without perturbation.

Although the dataset is not class-balanced, it reflects a realistic evaluation setting in which genuine references are more frequent than fabricated ones.
This composition also allows us to evaluate whether citation verification systems can detect hallucinated references without over-rejecting valid scholarly citations.
The controlled distribution across hallucination types ensures that evaluation results are not dominated by a single error mode and supports targeted analysis of model robustness across different classes of citation hallucination.

\subsection{Quality Control and Validation}

All generated hallucinated citations are subjected to retrieval-based checks followed by human validation.
A hallucinated citation is retained in the benchmark only if no authoritative scholarly record can be found that matches the perturbed metadata under standard normalization rules.
This process reduces the risk of labeling a valid but hard-to-index citation as hallucinated.

Through this validation procedure, the generated dataset preserves realistic citation form while ensuring that each hallucinated instance contains a verifiable bibliographic inconsistency.
As a result, the benchmark evaluates substantive citation verification ability rather than sensitivity to formatting artifacts or incomplete citation style conventions.

\section{Human Annotation and Verification Protocol}
\label{sec:annotation}

To ensure the reliability and correctness of benchmark labels, all citation annotations in both the generated and real-world datasets were conducted by the author team.
No crowd-sourced annotators were used, and no labels were assigned solely by automated systems.

\subsection{Annotation Scope}

Each citation instance is represented as a structured metadata tuple, including title, authors, venue, publication year, and persistent identifiers when available.
Annotators determine whether the citation corresponds to a valid scholarly record and whether its critical metadata fields are consistent with authoritative sources.

\subsection{Verification Procedure}

For each citation, annotators perform a manual verification process consisting of the following steps:
\begin{itemize}[left=0pt]
    \item \textbf{Authoritative Search:} The citation is queried against scholarly databases, publisher websites, Google Scholar, and institutional repositories.
    \item \textbf{Metadata Cross-Checking:} Retrieved records are manually compared against the citation title, author list, venue, year, and identifiers.
    \item \textbf{Existence Validation:} A citation is marked as \textit{Real} only if a corresponding scholarly work can be confidently identified with matching core metadata.
    \item \textbf{Hallucination Confirmation:} A citation is labeled as \textit{Hallucinated} if no matching scholarly record can be found or if critical metadata fields are demonstrably inconsistent.
\end{itemize}

\subsection{Labeling Criteria}

We adopt a strict but realistic labeling standard.
Minor formatting variations, capitalization differences, and missing optional fields such as page numbers are not considered hallucinations as long as the core bibliographic identity remains verifiable.
Conversely, inconsistencies in essential fields, including nonexistent titles, incorrect authorship, invalid venue attribution, or fabricated identifiers, result in a hallucinated label.

\subsection{Conflict Resolution}

All citation instances are independently reviewed by at least two authors.
In cases of disagreement or ambiguity, the citation is jointly re-examined by the author team until a consensus decision is reached.
Citations for which no consensus can be achieved after exhaustive verification are excluded from the benchmark to avoid label noise.

\subsection{Quality Assurance}

To further ensure annotation integrity, a subset of citations is randomly re-checked after the initial labeling process.
This auditing step helps prevent systematic bias and supports consistent application of the labeling criteria across the dataset.

\section{Prompt Templates and SOP Details}
\label{sec:prompts}

This appendix documents the core prompt templates used in our SOP-driven multi-agent citation verification framework.
For reproducibility and transparency, we disclose the prompts for the Planning Agent and the Judge Agent, which jointly control task routing and final verification decisions.

\subsection{Planning Agent Prompt}

The Planning Agent serves as the central SOP executor.
Its responsibility is not to judge citation correctness directly, but to determine the next verification stage for each citation.
It routes each citation through memory lookup, web verification, and scholar verification according to the predefined SOP.

\begin{promptbox}{Planning Agent Prompt}
You are a citation verification orchestrator.

Your task is to execute a strict Standardized Operating Procedure for academic citation verification.
You do not judge whether a citation is correct.
You only decide which verification step should run next.

Available stages:
\begin{enumerate}[leftmargin=*]
    \item \textbf{Memory Lookup:} Check whether the citation has already been verified.
    \item \textbf{Web Verification:} Compare the citation against evidence retrieved from web search.
    \item \textbf{Scholar Verification:} Use scholarly databases or authoritative bibliographic sources for final verification.
\end{enumerate}

SOP rules:
\begin{itemize}[leftmargin=*]
    \item Always attempt Memory Lookup first.
    \item If Memory Lookup confirms the citation, stop.
    \item If Memory Lookup fails or returns unknown, proceed to Web Verification.
    \item If Web Verification confirms a valid match, stop.
    \item If Web Verification fails or remains inconclusive, escalate to Scholar Verification.
    \item Scholar Verification is the final decision stage.
\end{itemize}

Constraints:
\begin{itemize}[leftmargin=*]
    \item Follow the SOP strictly.
    \item Do not skip stages.
    \item Do not modify citation metadata.
    \item Do not make the final correctness judgment yourself.
\end{itemize}

Return only a JSON object in the following format:

\begin{verbatim}
{
  "citation_id": "<id>",
  "next_action": "memory" | "web" | "scholar" | "stop",
  "reason": "<brief justification>"
}
\end{verbatim}
\end{promptbox}

The structured output enables deterministic execution and multi-threaded scheduling in the verification pipeline.

\subsection{Judge Agent Prompt}

The Judge Agent performs citation-to-evidence matching.
It receives citation metadata and retrieved search results, then determines whether at least one result supports the cited paper.
Unlike the Planning Agent, the Judge Agent makes the final verification decision, but it must base the decision only on retrieved evidence.

\begin{promptbox}{Judge Agent Prompt}
Decide whether the cited paper exists, given only the search results below.

Citation:
\begin{itemize}[leftmargin=*]
    \item Title: \{title\}
    \item Authors: \{authors\}
    \item Venue: \{venue\}
    \item Year: \{year\}
\end{itemize}

Search results:
\begin{verbatim}
{documents}
\end{verbatim}

Use only the provided search results.
Do not use outside knowledge.

\textbf{Title matching.}
First determine the source type.

For academic publications, including peer-reviewed papers, preprints, conference papers, journal articles, technical reports, theses, and books, apply strict title matching.
The title must match at the word level.
Only case, punctuation, whitespace, and hyphenation differences may be ignored.
Any added, removed, substituted, or reordered content word means the title does not match.

For non-academic online resources, including blog posts, documentation pages, dataset cards, model cards, GitHub pages, news articles, government documents, and standards pages, apply relaxed title matching.
Accept the title when it appears as a heading, project name, dataset name, model name, feature name, or a faithful description of the cited resource.
Reject it when the cited title has no meaningful presence on the resource.

\textbf{Author matching.}
Apply lenient author matching.
Author order does not matter.
Initials and full given names may match when surnames and initials are consistent.
Truncated author lists and ``et al.'' are acceptable when the overlapping surnames agree.
Transliteration variants, missing diacritics, and spacing or hyphenation differences in compound names are acceptable.

Reject on authors only when there is no meaningful surname overlap, or when a non-trivial subset of cited authors is clearly replaced by different authors.

\textbf{Venue and year.}
If the title matches and the authors agree, allow reasonable venue and year variation.
This includes arXiv versions and published versions of the same work, conference acronyms and full names, journal spelling or abbreviation variants, and year differences within two years for preprint-to-publication transitions.

\textbf{Evidence quality.}
Accept evidence from authoritative or bibliographic sources, including publisher pages, DOI pages, arXiv pages, ACL Anthology, PubMed, conference proceedings, Semantic Scholar, ResearchGate, and official institutional pages.
Reject cases where the only evidence is an unrelated paper bibliography that repeats the citation text without an independent publication record.

\textbf{Decision rule.}
Set \texttt{match=true} only when at least one search result contains a title match under the rules above and the authors agree under the lenient author rules.
Otherwise set \texttt{match=false}.

Return only JSON, with no markdown or extra explanation:

\begin{verbatim}
{
  "match": true | false,
  "matched_result": <result_number_or_null>,
  "note": "<short reason>"
}
\end{verbatim}
\end{promptbox}

\subsection{Role Separation and Determinism}

The Planning Agent and Judge Agent have strictly separated responsibilities.
The Planning Agent performs task routing only and never evaluates citation correctness.
The Judge Agent performs evidence-based matching under explicit rules and never controls pipeline execution.
This separation makes the verification process auditable and reduces the risk that routing decisions, retrieval failures, and final judgments become conflated.

\end{document}